\documentclass{article}
\pdfoutput=1

\usepackage{PRIMEarxiv}

\usepackage[utf8]{inputenc} 
\usepackage[T1]{fontenc}    
\usepackage{hyperref}       
\usepackage{url}            
\usepackage{booktabs}       
\usepackage{amsfonts}       
\usepackage{nicefrac}       
\usepackage{microtype}      
\usepackage{lipsum}
\usepackage{fancyhdr}       
\usepackage{graphicx}       
\graphicspath{{media/}}     

\title{Classification of Long Sequential Data using Circular Dilated Convolutional Neural Networks
}

\author{
  Lei Cheng, Ruslan Khalitov, Tong Yu, and Zhirong Yang\thanks{Corresponding author.} \\
  Department of Computer Science \\
  Norwegian University of Science and Technology \\
  \texttt{\{lei.cheng, ruslan.khalitov, tong.yu, zhirong.yang\}@ntnu.no} \\
}

\begin{document}
\maketitle

\begin{abstract}
Classification of long sequential data is an important Machine Learning task and appears in many application scenarios. Recurrent Neural Networks, Transformers, and Convolutional Neural Networks are three major techniques for learning from sequential data. Among these methods, Temporal Convolutional Networks (TCNs) which are scalable to very long sequences have achieved remarkable progress in time series regression. However, the performance of TCNs for sequence classification is not satisfactory because they use a skewed connection protocol and output classes at the last position. Such asymmetry restricts their performance for classification which depends on the whole sequence. In this work, we propose a symmetric multi-scale architecture called Circular Dilated Convolutional Neural Network (CDIL-CNN), where every position has an equal chance to receive information from other positions at the previous layers. Our model gives classification logits in all positions, and we can apply a simple ensemble learning to achieve a better decision. We have tested CDIL-CNN on various long sequential datasets. The experimental results show that our method has superior performance over many state-of-the-art approaches. The model and experiments are available at \url{https://github.com/LeiCheng-no/CDIL-CNN}.
\end{abstract}


\section{Introduction}
\label{sec:introduction}

Sequence classification is the task of predicting class labels for sequences. It is of central importance in many applications, such as document classification, genomic analysis, and health informatics. For example, classifying documents into different topic categories is a challenge for library science, especially for modern digital libraries \cite{khan2010review}. Genomic classification help researchers to further understand some diseases \cite{akbani2015genomic}. Classifying ECG time series tells if someone is a healthy person or a patient with heart disease \cite{led2004design}.

Machine Learning, especially Deep Learning, becomes widely used in end-to-end sequence classification, where a single model learns all steps between the initial inputs and the final outputs. Recurrent Neural Networks (RNNs), Transformers, and Convolutional Neural Networks (CNNs) are three primary techniques for analyzing sequential data.

RNNs use their internal states to process the sequence step by step. Despite success for short sequences, traditional RNNs cannot scale to very long sequences \cite{li2018independently}. One reason is that they are challenging to train due to exploding or vanishing gradient problems \cite{bengio1994learning}. In addition, the prediction of each timestep must wait for all its predecessors to complete, which makes RNNs difficult to parallelize. Transformers are a family of models relying on self-attention mechanism \cite{bahdanau2014neural, vaswani2017attention}. They have quadratic time and memory complexities to the input sequence length because they compute pairwise dot-products. Comprehensive approximations are required to reduce the cost \cite{tay2020efficient}.

In contrast, CNNs are able to handle very long sequences. A convolutional layer uses sparse connections and no recurrent nodes. Therefore, CNNs are easier to train and parallelize. In addition, dilated convolutions can exponentially enlarge the receptive fields, allowing CNNs to use fewer layers to capture long-term dependencies. For example, Temporal Convolutional Networks (TCNs) recently provide remarkable performance on sequence regression tasks \cite{bai2018empirical}. However, the performance of TCNs for classification tasks is not satisfactory. TCNs use causal convolutions which implement a skewed connection protocol. The asymmetric design causes a tendency to focus on the latter part of a sequence. 

In this paper, we propose a novel convolutional architecture named Circular Dilated Convolutional Neural Network (CDIL-CNN), which can scale to very long sequences and have superior performance on various classification tasks. Unlike TCNs, we use symmetric convolutions to mix information, and thus every position can receive both earlier and later information from previous layers in a circular manner. Unlike conventional pyramid-like CNN architecture, every position of the last convolutional layer in our design has an equal chance to receive all information from the whole sequence and gives its classification logits. Then a simple average ensemble learning helps our model achieve better accuracy.

We have tested our model on extensive sequence classification tasks, including synthetic data, images, texts, and audio series. Experimental results show that CDIL-CNN outperforms several state-of-the-art models. Our method can accurately and robustly classify across tasks with both short-term and long-term dependencies for very long sequences.

The remaining of the paper is organized as follows. We review some popular models for sequential data and their limitations in Section \ref{sec:related}. In Section \ref{sec:cdil}, we present our model, CDIL-CNN, including its connection protocol and network architecture. Experimental tasks and results are provided in Section \ref{sec:experiments}, and we discover that the simple convolutional network has superior performance over other models in various scenarios. Finally, we conclude the paper in Section \ref{sec:conclusion}.

\section{Related Works}
\label{sec:related}

A sequence $x$ of length $N$ is a list of elements $[x_1, x_2, \cdots, x_N]$, where $x_t\in\mathbb{R}^{D}~(1\leq t \leq N)$ is the $D$-dimensional element at the $t$-th position. Given a training set $\{x^{(i)}, y^{(i)}\}_{i=1}^I$ with $I$ sequences and their class labels, sequence classification uses the training set to fit a model $f:\mathbb{R}^{N \times D} \mapsto \mathbb{C}$, where $\mathbb{C}$ is the space of class labels. The fitted model can then be used to classify newly coming sequences.

Many deep neural networks have been proposed for various sequence classification tasks. RNNs, Transformers, and CNNs are three significant branches for learning from sequential data.

RNNs read and process inputs sequentially. At each timestep, an RNN takes the current sequence element and the hidden state as the input and outputs the next hidden state. The hidden state at a timestep is expected to act as the representation of all its earlier inputs. Because the prediction of each timestep must wait for all its predecessors to complete, the sequential process is difficult to parallelize, which makes RNNs hard to handle very long sequences. Moreover, basic RNNs suffer from vanishing and exploding gradient problems, making model training very difficult for long sequences \cite{bengio1994learning}. Gated RNNs, such as Long Short-Term Memory (LSTM) \cite{hochreiter1997long} and Gated Recurrent Unit (GRU) \cite{cho2014properties}, have been proposed to relieve the gradient problems. They have many additional gates to regulate the flow of information. The gated RNNs are used in many sequence classification tasks, such as ECG arrhythmia \cite{singh2018classification} and text \cite{sharfuddin2018deep, du2020novel}. However, they can process only short sequences (about 500-1000 timesteps) \cite{li2018independently}.

Transformers, a family of models based on attention mechanism, quantify the interdependence within the sequence elements (self-attention). Originally, attention was used in conjunction with recurrent networks and convolutional networks \cite{mnih2014recurrent, kim2017structured}. Later, Transformer, an architecture based solely on attention mechanism, was proposed. The vanilla Transformer computes pairwise dot-products between all sequence elements, which leads to a quadratic complexity w.r.t. the sequence length and makes it infeasible to process very long sequences. Approximated attention methods have been proposed to tackle this problem. Sparse Transformer \cite{child2019generating}, LogSparse Transformer \cite{li2019enhancing}, Longformer \cite{beltagy2020longformer}, and Big Bird \cite{zaheer2020big} use sparse attention mechanism. Linformer \cite{wang2020linformer} and Synthesizer \cite{tay2021synthesizer} apply low-rank projection attention. Performer \cite{choromanski2020rethinking}, Linear Transformer \cite{katharopoulos2020transformers}, and Random Feature Attention \cite{peng2021random} rely on kernel approximation. Reformer \cite{kitaev2020reformer}, Routing Transformer \cite{roy2021efficient}, and Sinkhorn Transformer \cite{tay2020sparse} follow the paradigm of re-arranging sequences. However, their approximation quality is questionable. Later in Section \ref{sec:experiments}, we will show that their performance is inferior for long sequence classification.  

CNNs are good at processing data that has a grid-like topology. Two-dimensional CNNs achieve great success in computer vision \cite{lecun1998gradient, krizhevsky2012imagenet, simonyan2014very, Szegedy_2015_CVPR}, while one-dimensional CNNs are commonly used for sequential data \cite{oord2016wavenet, kalchbrenner2016neural, gehring2016convolutional}. Among these models, TCNs which use causal convolutions with skewed connections attempt to capture the temporal interactions and have been applied to various regression tasks, such as action segmentation and detection \cite{lea2016temporal, lea2017temporal}, lip-reading \cite{martinez2020lipreading, ma2021lip}, and ENSO prediction \cite{yan2020temporal}. The comparison of the convolutional and recurrent architectures shows that a simple TCN outperforms canonical RNNs across a wide range of sequence modeling tasks \cite{bai2018empirical}.

\section{Circular Dilated CNN}
\label{sec:cdil}
Although TCN is suitable for long sequence regression, their performance for classification is not satisfactory. In this paper, we propose a new convolutional model, named CDIL-CNN, to overcome the TCN drawbacks in long sequence classification. More details are described as follows.

\subsection{Symmetric Dilated Convolutions}
\label{subsec:conv}

Our model uses symmetric convolutions that can receive both earlier and later information from previous layers. Because no information is allowed to be leaked from future to past in regression tasks, TCN uses causal convolutions that implement a skewed connection protocol, meaning that the output at timestep $t$ can only receive information of $t$ and earlier from previous layers. However, classification tasks do not have the restriction because the classification result depends on the whole sequence. Therefore, symmetric convolutions help our model better capture interactions.

Our model also uses increasing dilation sizes with the depth of the network. Dilated convolutions (or atrous convolutions) were originally introduced for dense image prediction, where they helped the model to capture multi-scale information \cite{yu2015multi, chen2017deeplab, chen2017rethinking, chen2018searching, wang2018understanding}. For 1D CNNs, dilated convolutions are generally used to enlarge the receptive fields \cite{oord2016wavenet, kalchbrenner2016neural, lea2017temporal, bai2018empirical}. Following these works, we increase the dilation sizes exponentially, i.e., $d_l=2^{l-1}$ where $d_l$ is the dilation size at the $l$-th convolutional layer. The combination of deep networks and exponentially dilated convolutions enables the receptive fields to expand quickly, which makes our model scalable to very long sequences. Our model needs $\lceil \log_2 \frac{N}{2} \rceil$ or $O(\log_2 N)$ layers to achieve a full receptive field for sequence length $N$.

\begin{figure}[tb]
	\centering
	\includegraphics[width=0.95\textwidth]{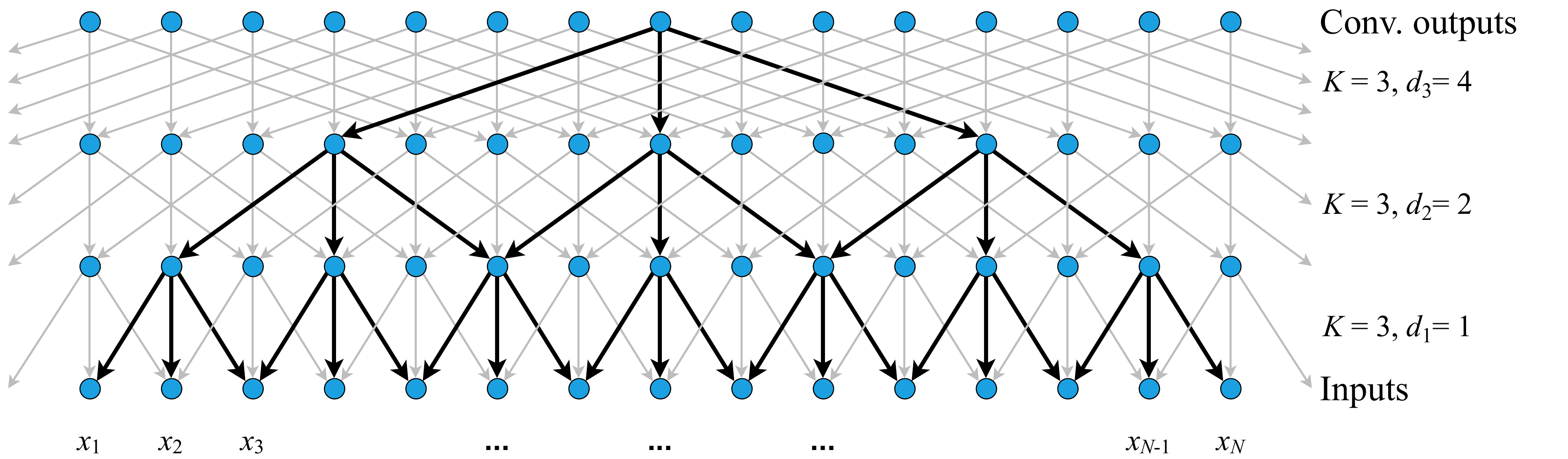}
	\caption{Illustration of symmetric dilated convolutions. Blue nodes represent the sequence elements. Each layer keeps the same size as the input sequence. Symmetric convolutions of kernel size 3 are used in all layers. Dilation sizes are increased exponentially.}
	\label{fig:dil}
\end{figure}

To avoid notional clutter, we start from the $D=1$ case. Let $[a_1, a_2, \cdots, a_N]$ denotes an 1-dimensional input sequence of the $l$-th convolutional layer. The convolutional output $b_t$ at the $t$-th ($1 \leq t \leq N$) position is computed by
$b_t = \sum\limits_{k=0}^{K-1} w^{(l)}_k \cdot a_{t+\left( k-\frac{K-1}{2} \right) \cdot d_l}$,
where the kernel size $K$ is usually an odd number\footnote{We used $K=3$ in all our experiments.} and $w^{(l)}$ are the convolution coefficients of the $l$-th layer. See Figure \ref{fig:dil} for an illustration of a 3-layer symmetric dilated convolutions with $K=3$. It is straightforward to extend the convolution with the bias term and for the $D>1$ cases.

\subsection{Circular Mixing}
\label{subsec:circular}

In traditional CNNs, zero-padding is often used for the boundary positions where the subscripts of their convoluted input positions $\left[ t+( k-\frac{K_l-1}{2} ) \cdot d_l \right]$ are smaller than 1 or larger than $N$. However, this can cause boundary effect because signals near the boundaries have to be mixed up with zeros and thus have less chance to be forwarded. The boundary effect creates blind spots and makes CNNs sensitive to absolute positions \cite{kayhan2020translation, alsallakh2020mind}. For example, CNNs with zero-padding can fail to capture the useful patterns if translation exists in the test data but not in the training data (see Section \ref{subsec:ablation}).

\begin{figure}[tb]
	\centering
	\begin{tabular}{ccc}
     \includegraphics[width=0.3\textwidth]{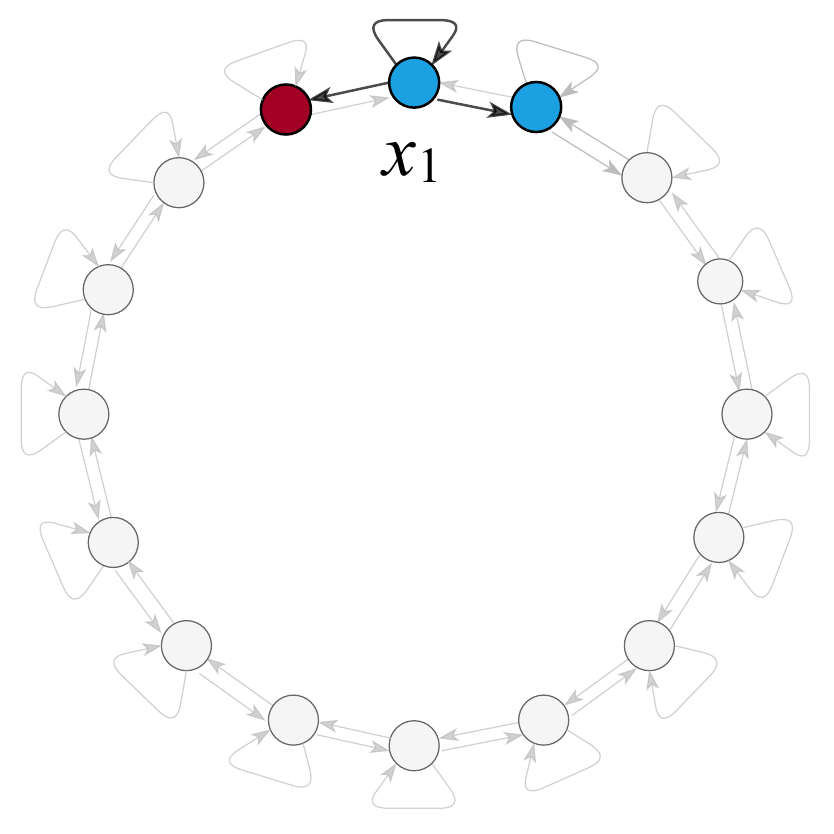}
    &\includegraphics[width=0.3\textwidth]{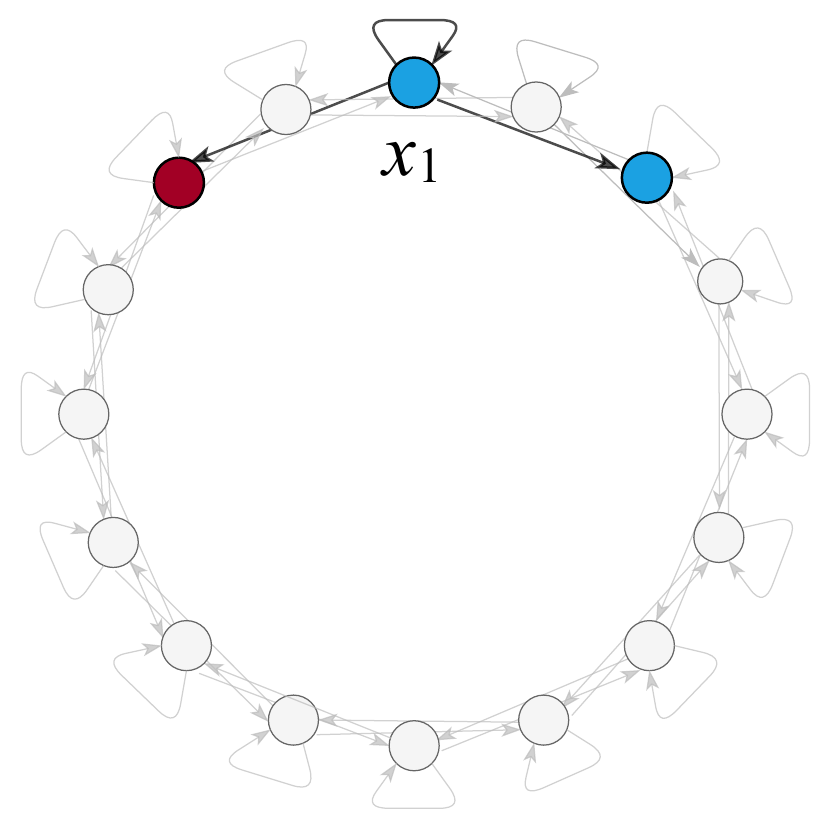}
    &\includegraphics[width=0.3\textwidth]{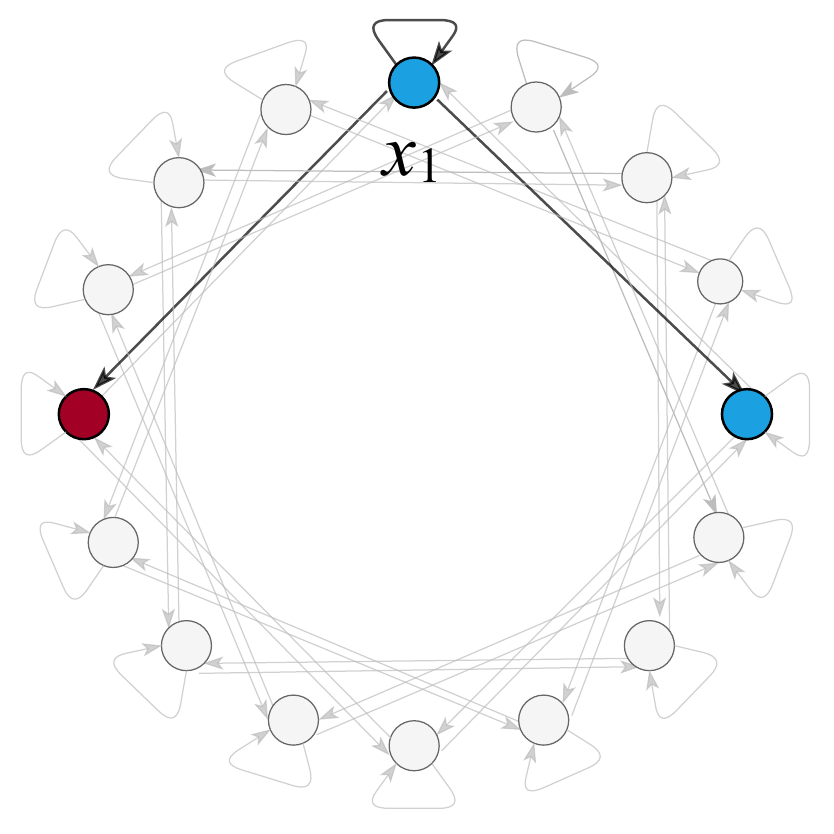}\\
    $K=3, d_1=1$, &$K=3, d_2=2$, &$K=3, d_3=4$\\
    \end{tabular}
	\caption{Illustrations of circular mixing. (a), (b), and (c) are the first, second, and third convolutional layer, respectively. Red nodes represent convoluted positions from the other end.}
	\label{fig:cir}
\end{figure}

We use a circular protocol because its corresponding circular padding can relieve the boundary effect \cite{kayhan2020translation, alsallakh2020mind}. In our model, a signal on one end is no longer convoluted with zeros but with signals from the other end.
Circular padding makes our model more robust to data shift and less sensitive to absolute position information.
The circular dilated convolutions are shown in Figure \ref{fig:cir}. The convolutional output $b_t$ becomes
\begin{equation}
    \label{equ:cdil}
    b_t = \sum\limits_{k=0}^{K-1} w^{(l)}_k \cdot a_{\left[ t+\left( k-\frac{K-1}{2} \right) \cdot d_l \right] \bmod N}
\end{equation}
Using circular dilated convolutions, our model can connect boundary positions and learn long-term dependencies even in the first layer, unlike lower layers of traditional CNNs which only focus on local information. In our design, every position of the last convolutional layer has an equal chance to receive all information of the whole input sequence. Therefore, our model can apply a simple average ensemble learning as below.

\subsection{Ensemble Learning}
\label{subsec:ensem}

We use a simple average ensemble learning to achieve better performance. RNNs and TCN assume that the last position contains all information of the whole sequence and the class decision depends only on the last position. In our model, every position of the last convolutional layer can receive all information of the whole sequence. A linear module $\mathbb{R}^{C} \mapsto \mathbb{C}$, where $C$ is the number of convolution channels, is applied on each convolutional output position, and each position gives its preliminary class logits. Then a simple average pooling as ensemble learning aggregates the individual logits. In the implementation, we can perform the average first to speed up the network because the linear module and the average pooling are exchangeable.

\begin{figure}[tb]
	\centering
	\includegraphics[width=0.35\textwidth]{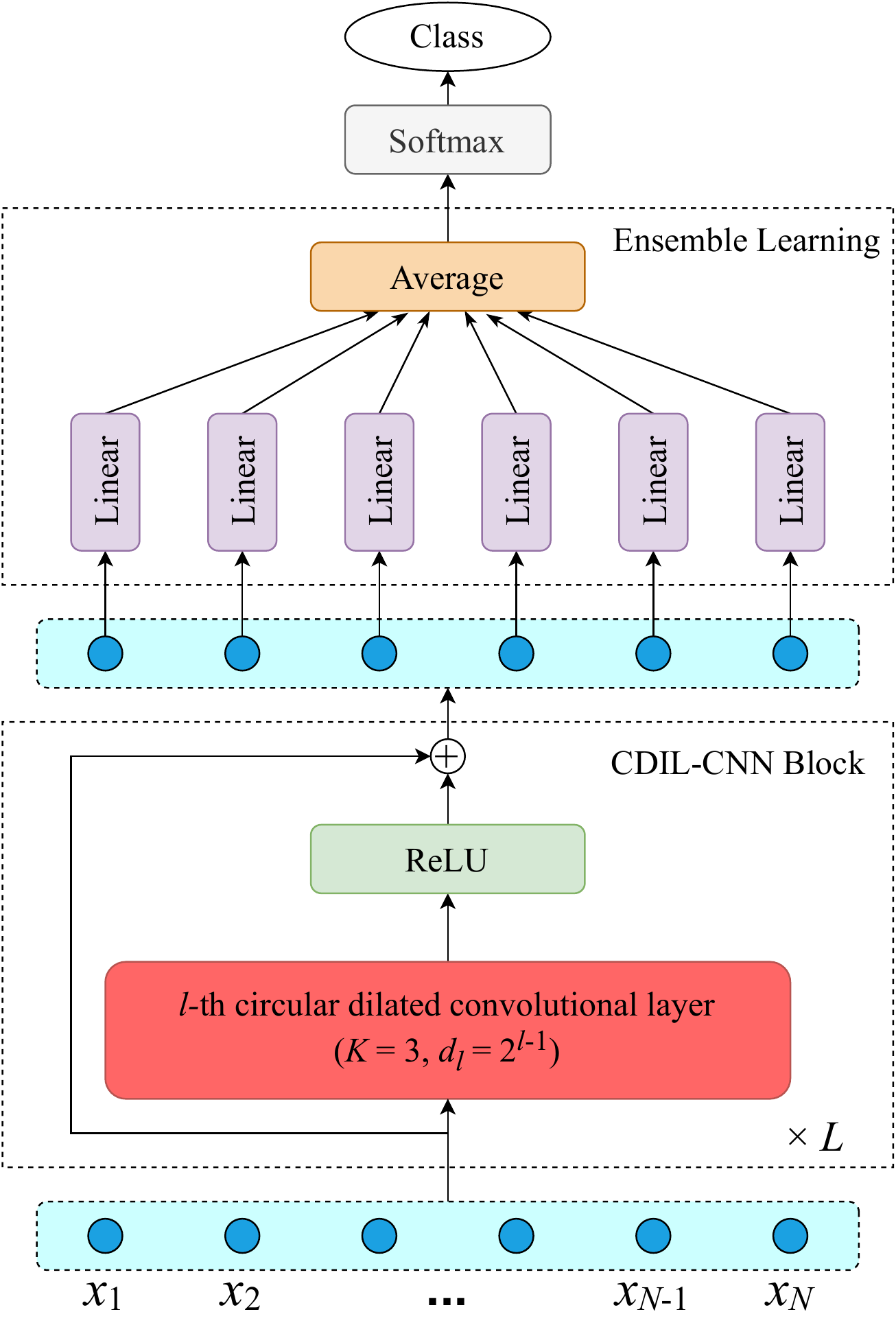}
	\caption{Our neural network architecture for sequence classification. The model comprises $L$ blocks of CDIL-CNN and an average ensemble learning. Each block outputs the same size as the input sequence. After the convolutions and the linear transformation, each position gives its prediction logits, and the average ensemble learning aggregates the logits.}
	\label{fig:res}
\end{figure}

Our model also uses residual connections to facilitate the training and to improve the accuracy \cite{he2016deep, he2016identity}. A residual block contains a skip connection where the inputs are added before the block outputs. A schematic view of our model is depicted in Figure \ref{fig:res}.

\section{Experiments}
\label{sec:experiments}

We have compared our model with many popular models (including RNNs, Transformers, and CNNs) on various long sequential datasets in three groups of experiments. First, we used a synthetic dataset with increasing sequence lengths to show the scalability of our model. Then, we tested our model on the Long Range Arena (LRA) benchmark suite which contains different dependencies. Finally, we tried three time series classification datasets that contain important local information and much noise. All experiments were run on a Linux server with one NVIDIA-Tesla V100 GPU with 32 GB of memory. More details are given in the supplemental document.

\subsection{Synthetic Task: XOR Problem}
\label{subsec:syn}

\begin{figure}[tb]
	\centering
	\includegraphics[width=0.7\textwidth]{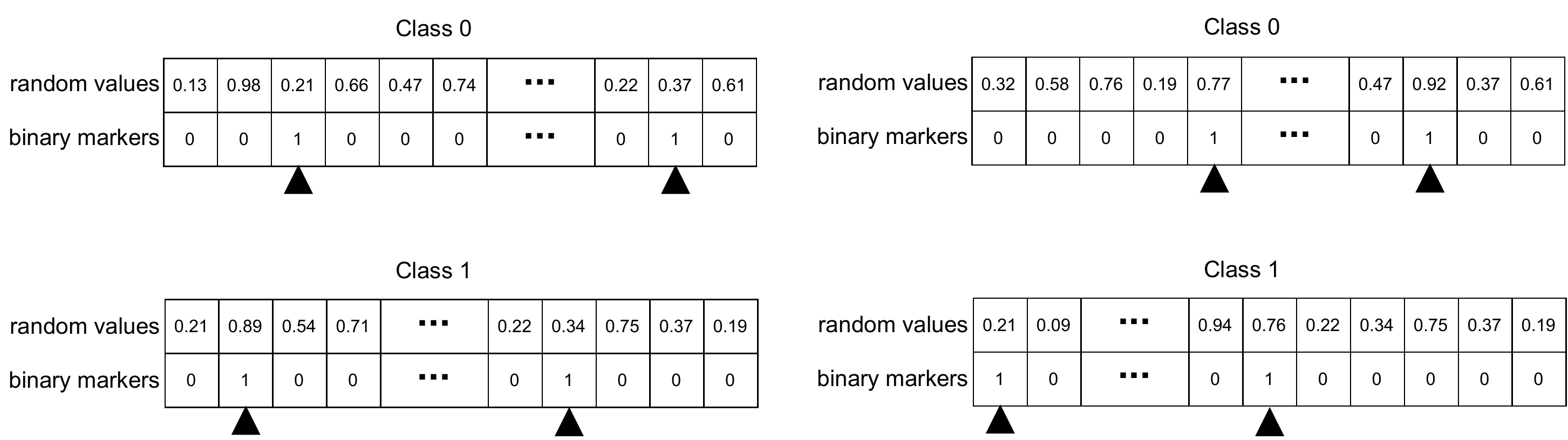}
	\caption{Examples of the XOR problem.}
	\label{fig:xor_example}
\end{figure}

The XOR problem is a classical classification problem in artificial neural network research which cannot be solved by a single perceptron \cite{minsky69perceptrons, rumelhart:errorpropnonote}. We created more challenging XOR tasks with increasing sequence lengths. For each length $N$, a sequence consists of $N$ pairs of numbers, where the first number, called value, is randomly chosen from the interval $[0, 1)$, and the second number is used as a marker. Most markers are 0 except two 1's at randomly selected positions. Let $X_1$ and $X_2$ denote the two values at the 1-marked positions. A sequence belongs to Class 0 if the values belong to the same half interval, i.e., $(X_1<0.5$ and $X_2<0.5)$ or $(X_1\geq0.5$ and $X_2\geq0.5)$. Otherwise, the sequence is labeled as Class 1. Figure \ref{fig:xor_example} shows four examples of the XOR problem. We have used $N=2^n$, where $n=4, \dots, 11$. A larger $N$ corresponds to a more challenging task. For each $N$, training, validation, and testing sets respectively have 10000 labeled sequences.

We have compared our model with several popular approaches: Transformer \cite{vaswani2017attention}, Linformer \cite{wang2020linformer}, Performer \cite{choromanski2020rethinking}, LSTM \cite{hochreiter1997long}, GRU \cite{cho2014properties}, TCN \cite{bai2018empirical}. We have also included
\begin{itemize}
    \item (Deformable): deformable convolutional networks that learn the adaptive receptive field using additional offsets \cite{dai2017deformable},
    \item (CNN): conventional convolutional neural networks with dilation size 1,
\end{itemize}
All convolutional networks use the $n-1$ layers and 32 channels for a fair comparison.

\begin{figure}[tb]
	\centering
	\includegraphics[width=0.7\textwidth]{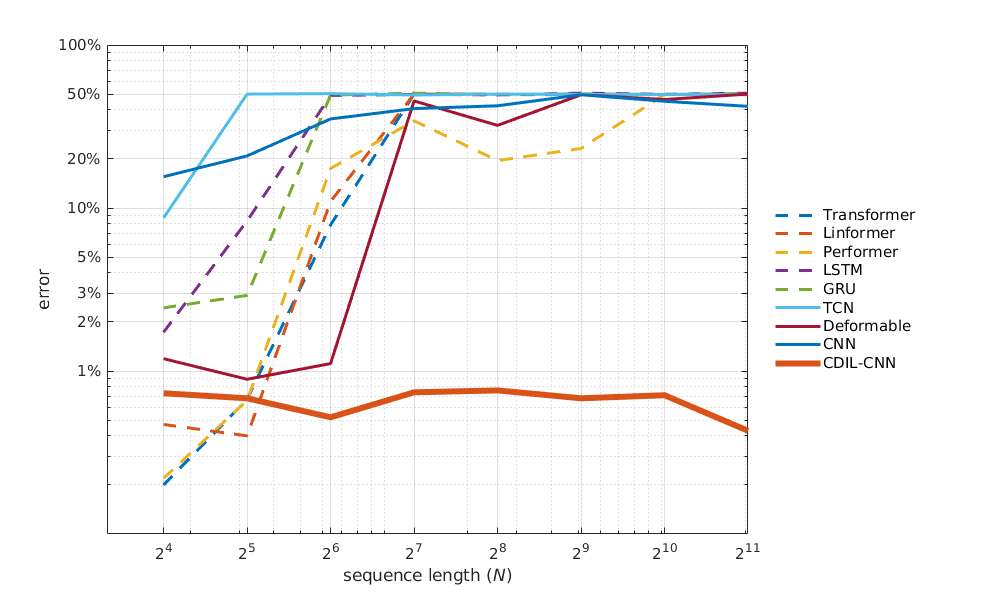}
	\caption{Error rate for the XOR problem with increasing sequence lengths.}
	\label{fig:xor}
\end{figure}

The results are shown in Figure \ref{fig:xor}. Our model performs accurately for all sequence lengths, where CDIL-CNN achieves less than 1\% error rate even when $N=2^{11}$. Transformer and its variants, RNNs, and Deformable achieve comparable error rates for short sequences. However, they turn inaccurate ($\sim50\%$ accuracy) when the sequences become longer than $N=128$. TCN and CNN perform even worse, where they respectively have 50\% and 20\% errors when $N=32$. The results indicate that CDIL-CNN is more scalable than the other compared methods.

\subsection{Long Range Arena Benchmark}
\label{subsec:lra}

\begin{figure}[tb]
	\centering
	\begin{tabular}{cccc}
            \includegraphics[width=0.2\textwidth]{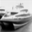} &
            \includegraphics[width=0.2\textwidth]{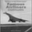} &
            \includegraphics[width=0.2\textwidth]{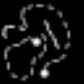} &
            \includegraphics[width=0.2\textwidth]{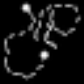} \\
            {\footnotesize Image, ship} &
            {\footnotesize Image, airplane} &
            {\footnotesize Pathfinder, negative} &
            {\footnotesize Pathfinder, positive}
    \end{tabular}
	\caption{Examples of Image (left two) and Pathfinder (right two)}
	\label{fig:lra}
\end{figure}

Long Range Arena is a public benchmark suite for evaluating model quality in long-context scenarios \cite{tay2020long}. The suite consists of different data types, such as images and texts. Many Transformers have been evaluated on the suite \cite{peng2021random, tay2020long, zhu2021long, xiong2021nystr}. We compared our CDIL-CNN with other models on the following datasets:
\begin{itemize}
    \item \textbf{Image}. This is a 10-class image classification task. The images come from the gray-scale version of CIFAR-10 \cite{krizhevsky2009learning}, where pixel intensities (0-255) are treated as categorical values. Two example images and their labels are shown in Figure \ref{fig:lra}. Every image is flattened to a sequence of length $N=1024$. The task requires the model to learn the 2D spatial relations while using the 1D sequences.
    
    \item \textbf{Pathfinder}. This is a synthetic image task motivated by cognitive psychology \cite{houtkamp2010parallel, linsley2018learning}. The task requires the model to make a binary decision whether two highlighted points are connected by a dashed path. Two example images and their labels are shown in Figure \ref{fig:lra}. Similar to the Image task, every pathfinder image is flattened to a sequence of length $N=1024$ with an alphabet size of 256.
    
    \item \textbf{Text}. This is a binary sentiment classification task of predicting whether an IMDb movie review is positive or negative \cite{maas2011learning}. The task considers the character-level sequences which generate longer inputs and make the task more challenging. We use a fixed length $N=4000$ for every sequence, which is truncated or padded when necessary.
    
    \item \textbf{Retrieval}. This is a character-level task with the ACL Anthology Network dataset \cite{radev2013acl}.  The task requires the model to process a pair of documents and determine whether they have a common citation. Like the Text task, every document is truncated or padded to the sequence length of $4000$, making the total length $N=8000$ for the pair.
\end{itemize}

For a fair comparison, we followed the same data preprocessing and training/validation/testing splitting in \cite{tay2020long}. We quoted the results of Transformer and its variants from the literature and ran RNNs and CNNs for completeness. We used one layer with a hidden size of 128 for RNNs and 64 channels for CNNs. All experiments were run five times with different random seeds, where means and standard deviations are reported in Table \ref{table:lra}. We have used paired $t$-test at the significance level of 0.05 to verify whether CDIL-CNN is significantly different from RNNs or other CNNs.

\begin{table}[tb]
\centering
\caption{Classification accuracy (\%) of different models on LRA tasks. $N$ is the sequence length. The dash means the result is absent in the reference paper. Means and standard deviations are computed by 5 runs. $\bullet$ denotes significant difference, and $\circ$ denotes insignificance.}
\label{table:lra}
\renewcommand{\arraystretch}{1.2}
\begin{tabular}{lcccc}
\hline
\hline
Model    &Image        &Pathfinder    &Text       &Retrieval   \\
         &$N$=1024     &$N$=1024      &$N=4000$   &$N=8000$       \\
\hline
Transformer~\cite{tay2020long}          & 42.44     & 71.40    & 64.27     & 57.46      \\
Transformer~\cite{xiong2021nystr}       & 38.20     & 74.16    & 65.02     & 79.35      \\
Transformer~\cite{zhu2021long}          & -         & -        & 65.35     & 82.30     \\
\hline
Local Attention~\cite{tay2020long}      & 41.46     & 66.63    & 52.98     & 53.39      \\
\hline
Sparse Transformer~\cite{tay2020long}    & 44.24     & 71.71    & 63.58     & 59.59      \\
\hline
Longformer~\cite{tay2020long}           & 42.22     & 69.71    & 62.85     & 56.89      \\
\hline
Linformer~\cite{tay2020long}            & 38.56     & 76.34    & 53.94     & 52.27      \\
Linformer~\cite{xiong2021nystr}         & 37.84     & 67.60    & 55.91     & 79.37      \\
Linformer~\cite{zhu2021long}            & -         & -        & 56.12     & 79.37     \\
\hline
Reformer~\cite{tay2020long}             & 38.07     & 68.50    & 56.10     & 53.40      \\
Reformer~\cite{xiong2021nystr}          & 43.29     & 69.36    & 64.88     & 78.64      \\
Reformer~\cite{zhu2021long}             & -         & -        & 64.88     & 78.64     \\
\hline
Sinkhorn Transformer~\cite{tay2020long}  & 41.23     & 67.45    & 61.20     & 53.83      \\
\hline
Synthesizer~\cite{tay2020long}          & 41.61     & 69.45    & 61.68     & 54.67      \\
\hline
BigBird~\cite{tay2020long}              & 40.83     & 74.87    & 64.02     & 59.29      \\
\hline
Linear Transformer~\cite{tay2020long}    & 42.34     & 75.30    & 65.90     & 53.09      \\
\hline
Performer~\cite{tay2020long}            & 42.77     & 77.05    & 65.40     & 53.82      \\
Performer~\cite{xiong2021nystr}         & 37.07     & 69.87    & 63.81     & 78.62      \\
Performer~\cite{zhu2021long}            & -         & -        & 65.21     & 81.70     \\
\hline
Nystr\"omformer~\cite{xiong2021nystr}   & 41.58     & 70.94     & 65.52     & 79.56     \\
Nystr\"omformer~\cite{zhu2021long}     & -         & -         & 65.75     & 81.29     \\
\hline
RFA-Gaussian~\cite{peng2021random}     & -         & -         & 66.0      & 56.1     \\
\hline
Transformer-LS~\cite{zhu2021long}      & -         & -         & 68.40     & 81.95     \\
\hline
LSTM        & $32.99\pm5.46\bullet$      & $61.26\pm12.14\bullet$    & $85.80\pm0.31\bullet$     & $77.18\pm0.23\bullet$  \\
\hline
GRU         & $44.40\pm1.12\bullet$      & $85.45\pm0.16\bullet$     & $86.70\pm0.21\bullet$     & $77.08\pm0.26\bullet$  \\
\hline
TCN         & $38.62\pm0.41\bullet$      & $85.48\pm0.46\bullet$     & $60.54\pm0.44\bullet$     & $76.85\pm0.08\bullet$  \\
\hline
Deformable  & $36.57\pm3.03\bullet$      & $56.14\pm0.48\bullet$     & $86.91\pm0.22\bullet$     & $83.69\pm0.97\circ$  \\
\hline
CNN         & $35.85\pm0.62\bullet$      & $55.95\pm0.06\bullet$     & $87.29\pm0.13\circ$     & $83.33\pm1.59\circ$  \\
\hline
CDIL-CNN    & $\textbf{64.49}\pm0.61$  & $\textbf{91.00}\pm0.37$  & $\textbf{87.61}\pm0.33$  & $\textbf{84.27}\pm0.76$   \\
\hline
\hline
\end{tabular}
\end{table}

Our model achieves the best mean accuracies in all tasks and is significantly better than RNNs and other CNNs in 17 out of 20 comparisons. The significant wins over all other methods hold for the Image and Pathfinder tasks. Espeicially for the Image task, CDIL-CNN achieves substantially higher mean accuracies (20.25\% better than the best transformer variant, 20.09\% better than the best RNN, 25.87\% better than other CNNs). Deformable and CNN get comparable accuracies with CDIL-CNN for Text and Retrieval, probably because the two tasks mainly rely on local patterns.

\subsection{Time Series}
\label{subsec:time}

The UEA \& UCR Repository\footnote{http://www.timeseriesclassification.com/} consists of various time series classification datasets \cite{dau2019ucr}. Many time series classification problems can be solved by detecting local patterns \cite{geurts2001pattern, ye2009time, iwana2020time}. These tasks require the model to pick out important local information from long sequences which contain much noise. We compared our CDIL-CNN with other popular models on three audio datasets:
\begin{itemize}
    \item \textbf{FruitFlies}. The dataset comes from the same optical sensor which recorded the change in amplitude of an infra-red light as it was occluded by the wings of fruit flies during flight. The dataset contains 17259 training and 17259 testing sequences of length $N=5000$. The task requires the model to classify a sequence as one of three species of the fruit fly.
    
    \item \textbf{RightWhaleCalls}. Right whale calls are difficult to hear due to some low-frequency anthropogenic sounds. Up-calls are the most commonly documented right whale vocalization. The task requires the model to decide whether a sequence contains a set of right whale up-calls or not. The training and testing sizes of this dataset are 10934 and 1962, respectively. All sequences have a fixed length $N=4000$.
    
    \item \textbf{MosquitoSound}. The dataset represents the wing beat of the flying mosquito. Both training and testing sets have 139883 instances with sequence length $N=3750$. The task requires the model to classify each sequence into one of six species.
\end{itemize}
We split every original training set into training (70\%) and validation (30\%) parts, and used the original testing set for testing.

We have compared our model with Transfomer, its two popular variants, RNNs, and CNNs. We also included dynamic convolutional neural networks (DCNNs) \cite{10.1145/3227609.3227690, 10.1007/978-3-030-88113-9_24}, because it combines CNN and dynamic time warping, a widely used component in many time series classifiers. We used 32 channels for every convolutional layer. The classification results are shown in Table \ref{table:time}.

\begin{table}[tb]
\centering
\caption{Classification accuracy (\%) of different models on time series datasets. $N$ is the sequence length. A DCNN run cannot finish in two days for the MosquitoSound dataset. Means and standard deviations are computed by 5 runs. All observed differences are statistically significant according to paired \textit{t}-test at the significance level (\textit{p}-value) of 0.05.}
\label{table:time}
\renewcommand{\arraystretch}{1.2}
\begin{tabular}{lccc}
\hline
\hline
Model       &FruitFlies   &RightWhaleCalls    &MosquitoSound   \\
            &$N=5000$     &$N=4000$           &$N=3750$       \\
\hline
Transformer     &$55.26\pm1.47$      &$71.84\pm0.65$       &$32.92\pm0.69$       \\
\hline
Linformer       &$81.80\pm1.61 $       &$71.17\pm0.84$       &$60.44\pm0.70$       \\
\hline
Performer       &$86.57\pm0.98 $      &$73.57\pm0.44$       &$68.34\pm0.88$       \\
\hline
LSTM            &$56.61\pm2.50$       &$61.39\pm6.61$       &$32.40\pm1.10$       \\
\hline
GRU             &$61.47\pm12.35$     &$63.18\pm8.54$       &$42.44\pm5.66$       \\
\hline
TCN             &$91.65\pm0.74$      &$86.92\pm0.38$       &$85.99\pm0.28$       \\
\hline
Deformable      &$92.68\pm1.70$       &$82.70\pm1.24$        &$88.92\pm0.43$       \\
\hline
DCNN            &$86.15\pm4.27$      &$69.98\pm1.58$       & -       \\
\hline
CNN             &$95.30\pm0.27$       &$78.34\pm1.05$       &$89.72\pm0.12$       \\
\hline
CDIL-CNN        &$\textbf{97.09}\pm0.08$    &$\textbf{91.99}\pm0.16$    &$\textbf{91.54}\pm0.22$      \\
\hline
\hline
\end{tabular}
\end{table}

Our model significantly wins all three tasks with mean accuracies of 97.09\%, 91.99\%, and 91.54\%, respectively. Transformers and RNNs struggle in the time series classification tasks. We found that convolutional networks perform better, probably because local signals are more important in these tasks. However, other CNNs are still inferior to our model.

\subsection{Ablation Study}
\label{subsec:ablation}

Compared with conventional CNN, the proposed CDIL-CNN has two major contributed components: dilated convolution and circular mixing (padding). In this section we performed an ablation study to verify that both components are conducive to accurate and robust classifications. For this goal, we include a middle method called DIL that contains only the dilated convolution component but zero-padding. We then compare DIL with conventional CNN and CDIL-CNN.

\begin{figure}[tb]
	\centering
	\includegraphics[width=0.95\textwidth]{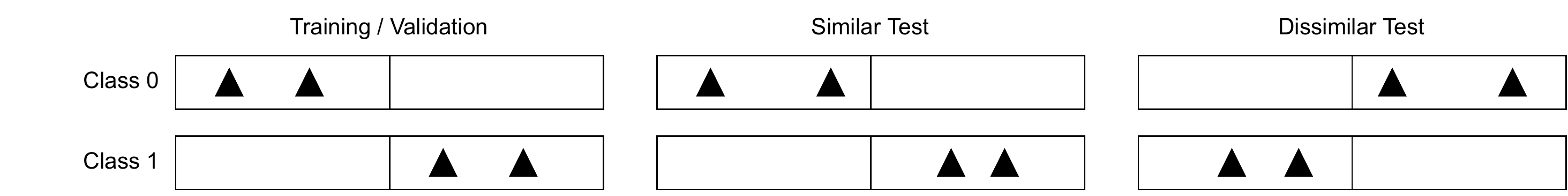}
	\caption{Position distribution of Similar Test and Dissimilar Test. Similar Test has the same position distribution as the training and validation datasets. Dissimilar Test flips the position distribution.}
	\label{fig:dissimilar}
\end{figure}

For comparison, we first designed a more challenging XOR problem, where $N=2^{11}$ and the test data can have the same position distribution as the training/validation data (Similar Test) or a different distribution (Dissimilar Test). See Figure \ref{fig:dissimilar} for illustration. In training/validation datasets, the two marked values appear in the first half for Class 0 and in the second half for Class 1. The test data follows the same pattern in the Similar Test, while the halves flip in the Dissimilar Test. The data shift brings an extra challenge, where a non-robust model can wrongly classify the sequences by the absolute positions of the markers instead of the required XOR pattern from marked values.

\begin{table}[tb]
\centering
\caption{Classification accuracy (\%) of Similar Test and Dissimilar Test. Means and standard deviations are computed by 5 runs.}
\label{table:dissimilar}
\renewcommand{\arraystretch}{1.2}
\begin{tabular}{lcc}
\hline
\hline
Model       & Similar Test   & Dissimilar Test   \\
\hline
CNN         & $50.79\pm0.57$       & $50.46\pm0.54$       \\
\hline
DIL         & $99.99\pm0.01$       & $0.81\pm0.42$       \\
\hline
CDIL-CNN    & $99.18\pm0.22$       & $98.91\pm0.37$       \\
\hline
\hline
\end{tabular}
\end{table}

The results are shown in Table \ref{table:dissimilar}. The CNN predictions are as bad as random guessing on both test sets, probably because it cannot capture the long-range interaction between the marked positions. DIL, equipped with dilated convolution, clearly improves the performance in Similar Test. However, DIL performs poorly on Dissimilar Test, which indicates that DIL overfits to training data and does not classify sequences by the required XOR pattern. CDIL-CNN differs from DIL by using circular padding instead of zero-padding. This change doesn't affect prediction performance in Similar Test, while achieves nearly perfect predictions in Dissimilar Test. The winning of CDIL-CNN shows that both dilated convolution and circular padding are needed for robust classification.

\begin{figure}[tb]
	\centering
	\begin{tabular}{cccc}
            \includegraphics[width=0.23\textwidth]{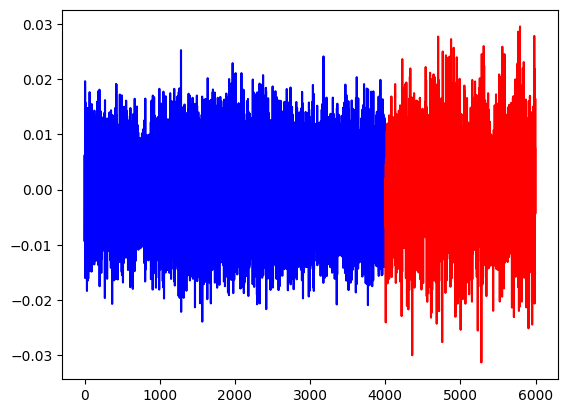} &
            \includegraphics[width=0.23\textwidth]{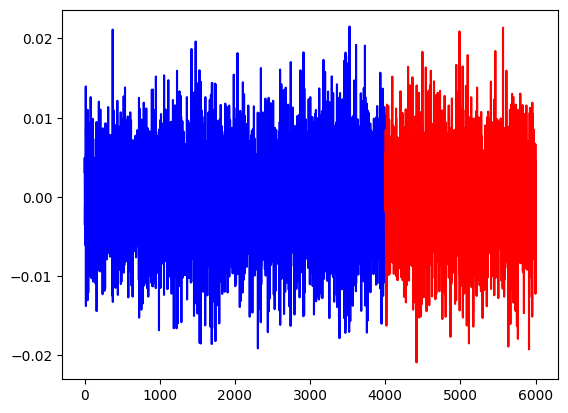} &
            \includegraphics[width=0.23\textwidth]{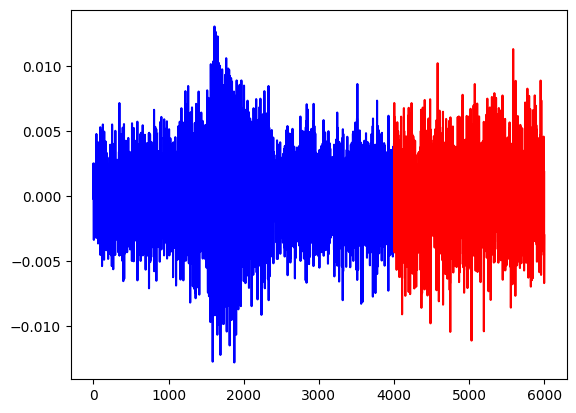} &
            \includegraphics[width=0.23\textwidth]{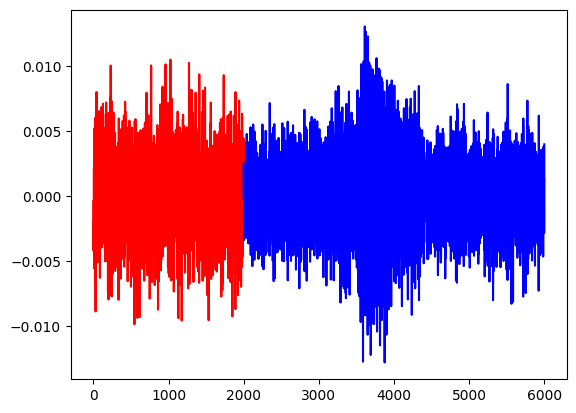} \\
            {\footnotesize Training} & 
            {\footnotesize Validation} & 
            {\footnotesize Similar Test} & 
            {\footnotesize Dissimilar Test} \\
    \end{tabular}
	\caption{Examples of noisy RightWhaleCalls. Blue is the original sequence, and red is the additional noise.}
	\label{fig:noise}
\end{figure}

We also created a noisy time series classification task using RightWhaleCalls, where the test data can have the same data shift as the training/validation data (Similar Test) or different shift (Dissimilar Test). We added the Gaussian noise of length 2000 at the end of every sequence in the training/validation set. The test set in Similar Test follows the same preprocessing, while in Dissimilar Test, the Gaussian noise part is inserted in front of each original test sequence. The mean and standard deviation of Gaussian noise equal those of the original sequence. Figure \ref{fig:noise} shows examples of noisy RightWhaleCalls.

The results are reported in Table \ref{table:noise}, which leads to similar conclusions in the XOR problem. CNN gives mediocre accuracies in both Similar Test and Dissimilar Test. Dilated convolution endows DIL better performance in Similar Test than CNN. However, zero-padding makes DIL sensitive to the data shift and degrades its performance close to random guessing in Dissimilar Test. Equipped with both dilated convolution and circular padding, CDIL-CNN can robustly and accurately classify (higher than 91\% accuracy) the time series in both cases.

\begin{table}[tb]
\centering
\caption{Classification accuracy (\%) of noisy RightWhaleCalls. Means and standard deviations are computed by 5 runs.}
\label{table:noise}
\renewcommand{\arraystretch}{1.2}
\begin{tabular}{lcc}
\hline
\hline
Model       & Similar Test   & Dissimilar Test   \\
\hline
CNN         & $78.20\pm0.65$       & $78.12\pm0.97$       \\
\hline
DIL         & $91.20\pm0.26$       & $50.33\pm3.87$       \\
\hline
CDIL-CNN    & $91.33\pm0.33$       & $91.39\pm0.37$       \\
\hline
\hline
\end{tabular}
\end{table}

\begin{figure}[tb]
	\centering
	\begin{tabular}{ccc}
        \includegraphics[width=0.45\textwidth]{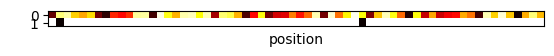} &
        \includegraphics[width=0.45\textwidth]{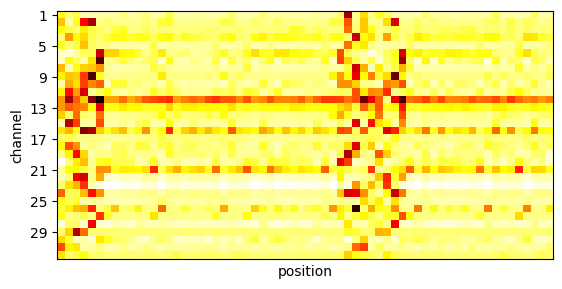} &\\
        {\footnotesize Input sequence} & 
        {\footnotesize Output features of CNN} & \\
        \includegraphics[width=0.45\textwidth]{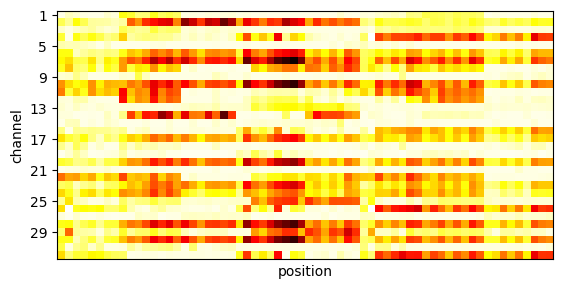} &
        \includegraphics[width=0.45\textwidth]{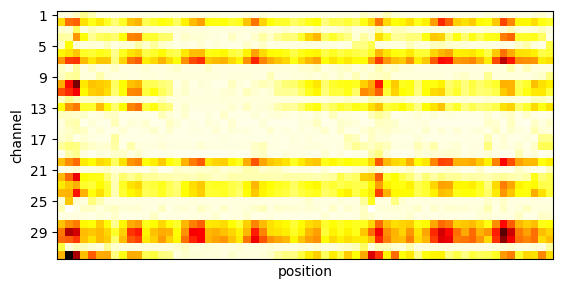} & 
        \includegraphics[width=0.05\textwidth]{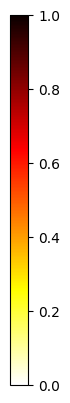} \\
        {\footnotesize Output features of DIL} & 
        {\footnotesize Output features of CDIL-CNN}& \\
    \end{tabular}
	\caption{Normalized matrix plots of an example input sequence and its outputs by the last layer of CNN, DIL, and CDIL-CNN. Darker colors indicate larger values.}
	\label{fig:feature}
\end{figure}

Next, we visualized an input sequence of the XOR problem and its output features of CNN, DIL, and CDIL-CNN in Figure \ref{fig:feature}. The visualization helps us understand the difference among the methods in terms of receptive field and boundary effect. In conventional CNN, the important information is present locally even in the last layer, which can lead to a wrong prediction if the two markers are distant. In contrast, the receptive field in DIL and CDIL-CNN is much wider because they use dilated convolution. It is known that zero-padding can cause boundary artifacts \cite{kayhan2020translation, alsallakh2020mind}. As we can see, here DIL has such artifacts in the left-most part of its visualization. Consequently, DIL probably misses the left marked value and thus gives wrong classification. In comparison, CDIL-CNN with circular padding leads to more even output features across columns and does not suffer from boundary artifacts.

In summary, both dilated convolution and circular padding are useful for robust and accurate classification. With the two components, CDIL-CNN well mixes the signals from the input sequence to every output position. As a result, the subsequent averaging and linear classifier provide a good ensemble and do not lessen the contributions of the important information.

\section{Conclusions}
\label{sec:conclusion}

We have proposed a novel convolutional model named Circular Dilated Convolutional Neural Network (CDIL-CNN) for sequence classification. Based on the characteristic of very long sequential data, we have used a design that consists of multiple symmetric and circular convolutions with exponential dilation sizes. Therefore, our model can remove boundary effect and enlarge the receptive fields quickly. In this way, every position of the last convolutional layer has an equal chance to receive all information of the whole input sequence. Finally, a simple average ensemble learning is applied to improve the accuracy. Experimental results show that our model has superior performance over all other models on various long sequential datasets.

In the future, we could add other popular modules to our model, such as absolute positional encoding \cite{vaswani2017attention}, relative positional encoding \cite{shaw2018self}, and conditional positional encoding \cite{chu2021conditional}, which could further improve the performance. We could also pre-train our model for few-shot or zero-shot learning, where only a few supervised labels are required in training.

\section{Acknowledgements}
We acknowledge for using the IDUN computing cluster \cite{idun}.

\bibliographystyle{unsrt}  
\bibliography{references}

\clearpage
\begin{center}
\textbf{\large Supplemental Document. Classification of Long Sequential Data using Circular Dilated Convolutional Neural Networks}
\end{center}

\setcounter{equation}{0}
\setcounter{figure}{0}
\setcounter{table}{0}
\setcounter{page}{1}
\setcounter{section}{0}
\section{XOR Problem}
In this group of experiments, we used the categorical cross-entropy loss function and the Adam optimizer~\cite{kingma2014adam} with the learning rate of 0.001. We trained every model for 100 epochs using the batch size of 40. For RNNs, namely LSTM and GRU, we used 1 layer with a hidden size of 128. For Transformer, Linformer, and Performer, we used 32 dimensions, 4 layers, and 4 heads. In CNNs, we used the kernel size of 3 and 32 channels for every convolutional layer. We adopted the varying depth of CNNs so that the last position of TCN and each position of CDIL-CNN can cover the whole sequence, i.e., the depth $L = n-1$ for the sequence length $N=2^n$. Other convolutional networks used the same layers for a fair comparison. Table~\ref{table:app_syn} gives the model sizes.

\begin{table}[bh]
\centering
\caption{The number of parameters ($\approx K$) for every model on the XOR Problem. $N$ is the sequence length. CNNs include TCN, CNN, DIL, and CDIL-CNN.}
\label{table:app_syn}
\renewcommand{\arraystretch}{1.2}
\begin{tabular}{lcccccccc}
\hline
\hline
Model       &$N=2^4$    &$N=2^5$    &$N=2^6$    &$N=2^7$    &$N=2^8$    &$N=2^9$     &$N=2^{10}$    &$N=2^{11}$\\
\hline
Transformer &26.02     &26.02     &26.02     &26.02     &26.02     &26.02     &26.02     &26.02     \\
\hline
Linformer   &45.47     &47.52     &51.62     &59.81     &76.19     &108.96    &174.50    &305.57    \\
\hline
Performer   &101.28    &101.28    &101.28    &101.28    &101.28    &101.28    &101.28    &101.28    \\
\hline
LSTM        &67.84     &67.84     &67.84     &67.84     &67.84     &67.84     &67.84     &67.84     \\
\hline
GRU         &50.95     &50.95     &50.95     &50.95     &50.95     &50.95     &50.95     &50.95     \\
\hline
CNNs        &6.69      &9.83      &12.96     &16.10     &19.23     &22.37     &25.51     &28.64     \\
\hline
Deformable  &8.30      &12.25     &15.39     &18.53     &21.66     &24.80     &27.93     &31.07     \\
\hline
\hline
\end{tabular}
\end{table}

\section{Long Range Arena}
For this group of experiments, we quoted Transformers' results from reference papers~\cite{peng2021random, tay2020long, zhu2021long, xiong2021nystr} and ran LSTM, GRU, TCN, CNN, Deformable, and CDIL-CNN for comparison. During training, we used the categorical cross-entropy loss function and the Adam optimizer with the learning rate of 0.001. For LSTM and GRU, we used 1 layer with a hidden size of 128. We used the kernel size of 3 and 64 channels for every convolutional layer. The depth was decided by the sequence length. All tasks had a vocabulary size of 256 and an embedding dimension of 64. Every model was trained for 100 epochs. More details of RNNs and CNNs are given in Table~\ref{table:app_lra}.

\begin{table}[bh]
\centering
\caption{Hyperparameters details of RNNs and CNNs for every LRA task. $N$, $C$, $B$, $L$, $P_M$ refer to the sequence length, classes, batch size, CNN depth, and parameter size ($\approx K$) of the model $M$, respectively. CNNs include TCN, CNN, and CDIL-CNN.}
\label{table:app_lra}
\renewcommand{\arraystretch}{1.2}
\begin{tabular}{lcccccccc}
\hline
\hline
Task         &$N$      &$C$      &$B$    &$P_{LSTM}$    &$P_{GRU}$    &$L$    &$P_{CNNs}$    &$P_{Deformable}$\\
\hline
Image        &1024     &10    &32     &117.00     &92.17     &9     &128.78     &133.61     \\
\hline
Pathfinder   &1024     &2     &256    &115.97     &91.14     &9     &128.26     &133.09     \\
\hline
Text         &4000     &2     &32     &115.97     &91.14     &11    &153.09     &157.92     \\
\hline
Retrieval    &8000     &2     &256    &116.74     &91.91     &11    &153.47     &158.30     \\
\hline
\hline
\end{tabular}
\end{table}

\section{Time Series}
In this group of experiments, we used the categorical cross-entropy loss function and the Adam optimizer with the learning rate of 0.001. We trained every model for 100 epochs using the batch size of 64. For LSTM and GRU, we used 1 layer with a hidden size of 128. For Transformer, Linformer, and Performer, we used 32 dimensions, 4 layers, and 4 heads. For CNNs, we used kernel size of 3 and 32 channels for every convolutional layer. The depth was decided by the sequence length. More details are given in Table~\ref{table:app_time}.

\begin{table}[h]
\centering
\caption{Hyperparameters details for every time series task. $L$ and $P_M$ refer to the CNN depth, and parameter size ($\approx K$) of the model $M$, respectively. CNNs include TCN, CNN, and CDIL-CNN.}
\label{table:app_time}
\renewcommand{\arraystretch}{1.2}
\begin{tabular}{lccccccccc}
\hline
\hline
Task              &$P_{Trans.}$    &$P_{Lin.}$    &$P_{Per.}$    &$P_{LSTM}$    &$P_{GRU}$    &$L$    &$P_{CNNs}$    &$P_{DCNN}$         &$P_{Deformable}$\\
\hline
FruitFlies        &26.02     &683.43     &101.28     &67.46     &50.69     &12     &34.82     &34.60     &37.25 \\
\hline
RightWhaleCalls   &25.99     &555.39     &101.25     &67.33     &50.56     &11     &31.65     &31.43     &34.08 \\
\hline
MosquitoSound     &26.12     &523.53     &101.38     &67.85     &51.08     &11     &31.78     & -     &34.21 \\
\hline
\hline
\end{tabular}
\end{table}

\end{document}